\documentclass[default,iicol,sn-aps]{sn-jnl}

\usepackage{graphicx} 
\usepackage{bm} 
\usepackage{amsmath} 
\usepackage{bbm} 
\usepackage{wrapfig} 
\usepackage{algorithm} 
\usepackage{algpseudocode} 
\usepackage{multirow} 
\usepackage{makecell}
\usepackage{booktabs}
\usepackage{colortbl}
\usepackage{color,xcolor}
\usepackage{multirow}

\jyear{2022}
\theoremstyle{thmstyleone}
\theoremstyle{thmstyletwo}

\theoremstyle{thmstylethree}

\graphicspath{{./figure/}}
\begin{document}
\title[Article Title]{DeepDR: an integrated deep-learning model web server for drug repositioning}
\author[1]{\fnm{Shuting} \sur{Jin}}\email{stjin@stu.xmu.edu.cn}
\author[2]{\fnm{Yi} \sur{Jiang}}\email{jiangyisoftware@mail.sdu.edu.cn}
\author[3]{\fnm{Yimin} \sur{Liu}}\email{yiminliu.career@gmail.com}
\author[4]{\fnm{Tengfei} \sur{Ma}}\email{tfma@hnu.edu.cn}
\author[5]{\fnm{Dongsheng} \sur{Cao}}\email{oriental-cds@163.com}
\author*[6]{\fnm{Leyi} \sur{Wei}}\email{weileyi@sdu.edu.cn}
\author*[7]{\fnm{Xiangrong} \sur{Liu}}\email{xrliu@xmu.edu.cn}
\author*[4]{\fnm{Xiangxiang} \sur{Zeng}}\email{xzeng@hnu.edu.cn}

\affil[1]{\orgdiv{School of Computer Science and Technology}, \orgname{Wuhan University of Science and Technology}, \orgaddress{\city{Wuhan}, \postcode{430065}, \country{China}}}
\affil[2]{\orgdiv{Department of Biomedical Informatics, College of Medicine}, \orgname{The Ohio State University}, \orgaddress{\city{Columbus, OH}, \postcode{43210}, \country{USA}}}
\affil[3]{\orgdiv{School of Life and Pharmaceutical Sciences}, \orgname{Dalian University of Technology}, \orgaddress{\city{Panjin},\country{China}}}
\affil[4]{\orgdiv{School of Information Science and Engineering}, \orgname{Hunan University}, \orgaddress{\city{Hunan}, \postcode{410082}, \country{China}}}
\affil[5]{\orgdiv{Xiangya School of Pharmaceutical Sciences}, \orgname{Central South University}, \orgaddress{\city{Changsha}, \postcode{410013}, \country{China}}}
\affil[6]{\orgdiv{Centre for Artificial Intelligence driven Drug Discovery, Faculty of Applied Science,}, \orgname{Macao Polytechnic University}, \orgaddress{\city{Macao SAR}, \country{China}}}
\affil[7]{\orgdiv{School of Informatics}, \orgname{Xiamen University}, \orgaddress{\city{Xiamen}, \postcode{361005}, \country{China}}}

\abstract{
\textbf{Background:} Identifying new indications for approved drugs is a complex and time-consuming process that requires extensive knowledge of pharmacology, clinical data, and advanced computational methods. Recently, deep learning (DL) methods have shown their capability for the accurate prediction of drug repositioning.  However, implementing DL-based modeling requires in-depth domain knowledge and proficient programming skills. \\
\textbf{Results:} In this application, we introduce {\bf DeepDR}, the first integrated platform that combines a variety of established DL-based models for disease- and target-specific drug repositioning tasks. {\bf DeepDR} leverages invaluable experience to recommend candidate drugs, which covers more than 15 networks and a comprehensive knowledge graph that includes 5.9 million edges across 107 types of relationships connecting drugs, diseases, proteins/genes, pathways, and expression from six existing databases and a large scientific corpus of 24 million PubMed publications. Additionally, the recommended results include detailed descriptions of the recommended drugs and visualize key patterns with interpretability through a knowledge graph.  \\
\textbf{Conclusion:} {\bf DeepDR} is free and open to all users without the requirement of registration. We believe it can provide an easy-to-use, systematic, highly accurate, and computationally automated platform for both experimental and computational scientists.
}

\maketitle
\section{Introduction}
Among the tens of thousands of diseases known to humans, although most diseases can be effectively treated, there are still many diseases in the dilemma of no medicines available. Drug repositioning aims to expand existing indications or discover new targets by studying the approved drugs, thereby offering a relatively low-cost and high-efficiency approach towards the rapid development of efficacious treatments \cite{padhy2011drug}. Repositioning already-approved drugs for new indications has become a promising tactic in drug development, and the global COVID-19 pandemic has further highlighted the urgent need for repurposing existing drugs to treat emerging infectious diseases promptly \cite{thanapalasingam2022relational}.  

Although there are many successful cases of drug repositioning, the discovery of new indications of approved drugs has been largely serendipitous. Systematic drug repositioning methods can be divided into two categories, one is the experimental method mainly based on high-throughput screening and high-content technology, and the other is the computational method based on computer-based virtual screening and prediction. However, due to cost constraints, the experimental methods can only screen a small number of drugs \cite{rudrapal2020drug}, and the screening results will be affected by factors such as the chemical properties and stability of the drug. 
In order to improve screening efficiency, most of the currently adopted strategies combine these two methods. First, \textit{in silico} methods are used for virtual screening to select candidate drugs with good druggability and development prospects, and then further screening and verification are carried out by experimental screening methods. 

Over the past few decades, computational techniques that incorporate information from genome-wide and context-specific molecular networks have achieved state-of-the-art results. Among the various computational methods that have been established for drug repositioning, deep learning (DL) methods have become the mainstay. DL has revolutionized traditional computational modeling of compounds by offering increased expressive power in identifying, processing, and extrapolating complex patterns in molecular data. Our recent researches \cite{pan2022deep,zeng2022accurate} have demonstrated that DL models can extract features from the topology of molecular networks, providing a reliable and data-driven method for predicting the association of a drug with a given input disease or target. We believe these approaches are a promising avenue for drug discovery and development. 

\begin{figure}[h]
\centering
\includegraphics[scale=0.32]{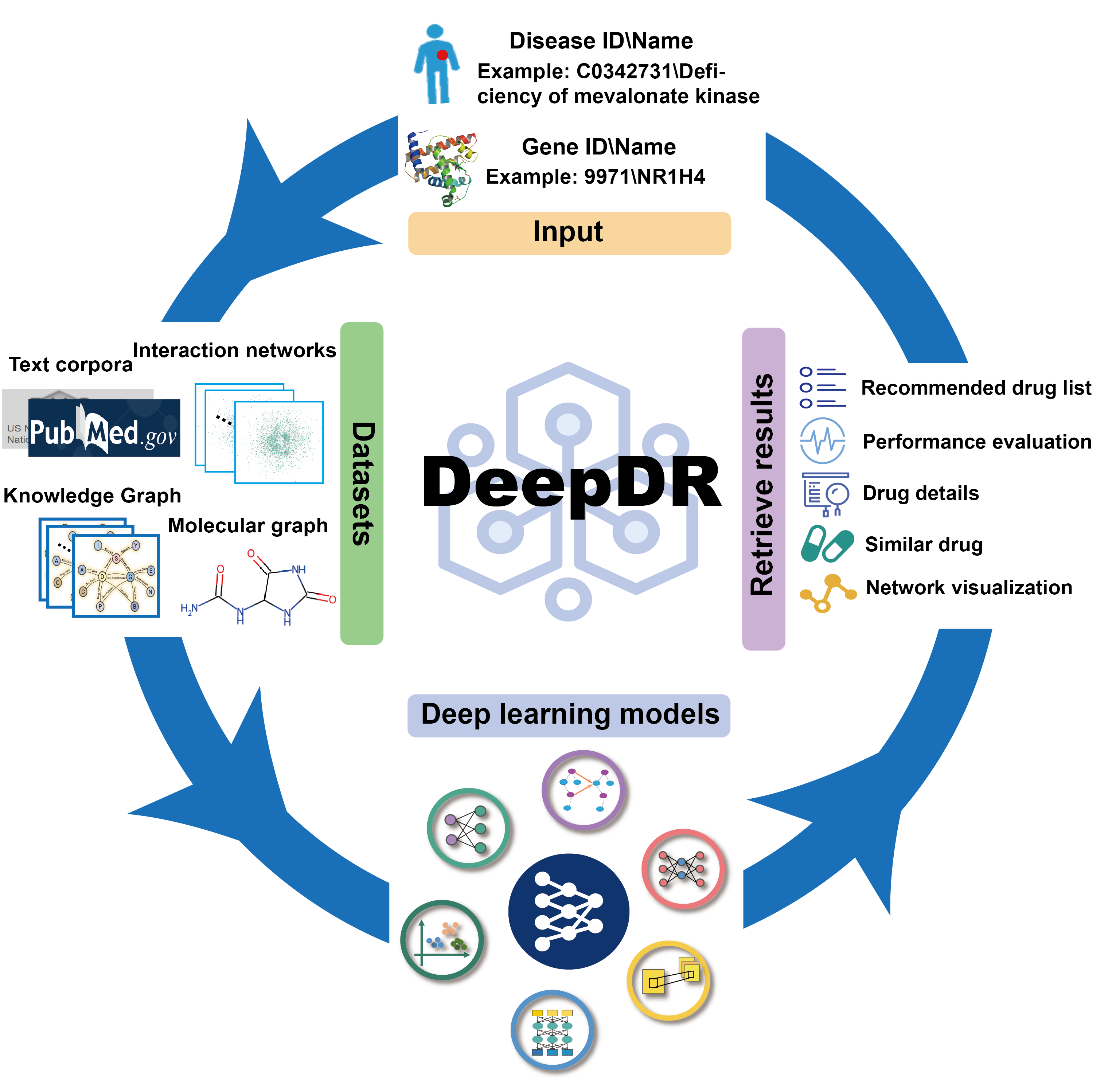}
\caption{\textbf{The graph abstract of {\bf DeepDR}.} }
\label{fig1}
\end{figure}

As powerful as these computational methods can be, their impact is fully realized only if they can be accessible to biomedical researchers, regardless of their programming and computational background \cite{huang2020deeppurpose}. A publicly available web server is an excellent platform for disseminating these results and an ideal web server would have the following properties:
\begin{itemize}
    \item Allow users to choose the most suitable model for biological problems from a set of deep learning models.
    \item Provide users with an intuitive and user-friendly interface, regardless of programming skills, and offer extensive help and tutorials.
    \item Offer users dependable predictive insights and present the detailed information of the prediction results through visualizations.
    \item Open-source data and code for predictive models and web servers are freely available.
\end{itemize}

To integrate multiple types of DL methods and assist bioinformaticians in accelerating the drug discovery process, we build {\bf DeepDR} (Deep-learning based Drug Repositioning predictor) to predict potential drugs for a given disease or target (see Figure \ref{fig1}). {\bf DeepDR} is a public web server for users without a coding background. A user can input a target (gene) or disease and choose the desired model, then the web server trains a custom supervised deep learning model using the known datasets. Within a few minutes, the user can retrieve top-ranked drugs for the input target or disease. It is worth noting that we visualize the prediction results and give detailed information of the top-ranked drugs. We also explain the prediction by visualizing key patterns of the top-ranked drugs from the trained KG. In the ``Resources" page of {\bf DeepDR}, users also can directly access and obtain source data resources. 

\section{{\bf DeepDR} platform} 
\subsection{The overall framework of {\bf DeepDR}}
\begin{figure*}[h]
\centering
\includegraphics[scale=0.28]{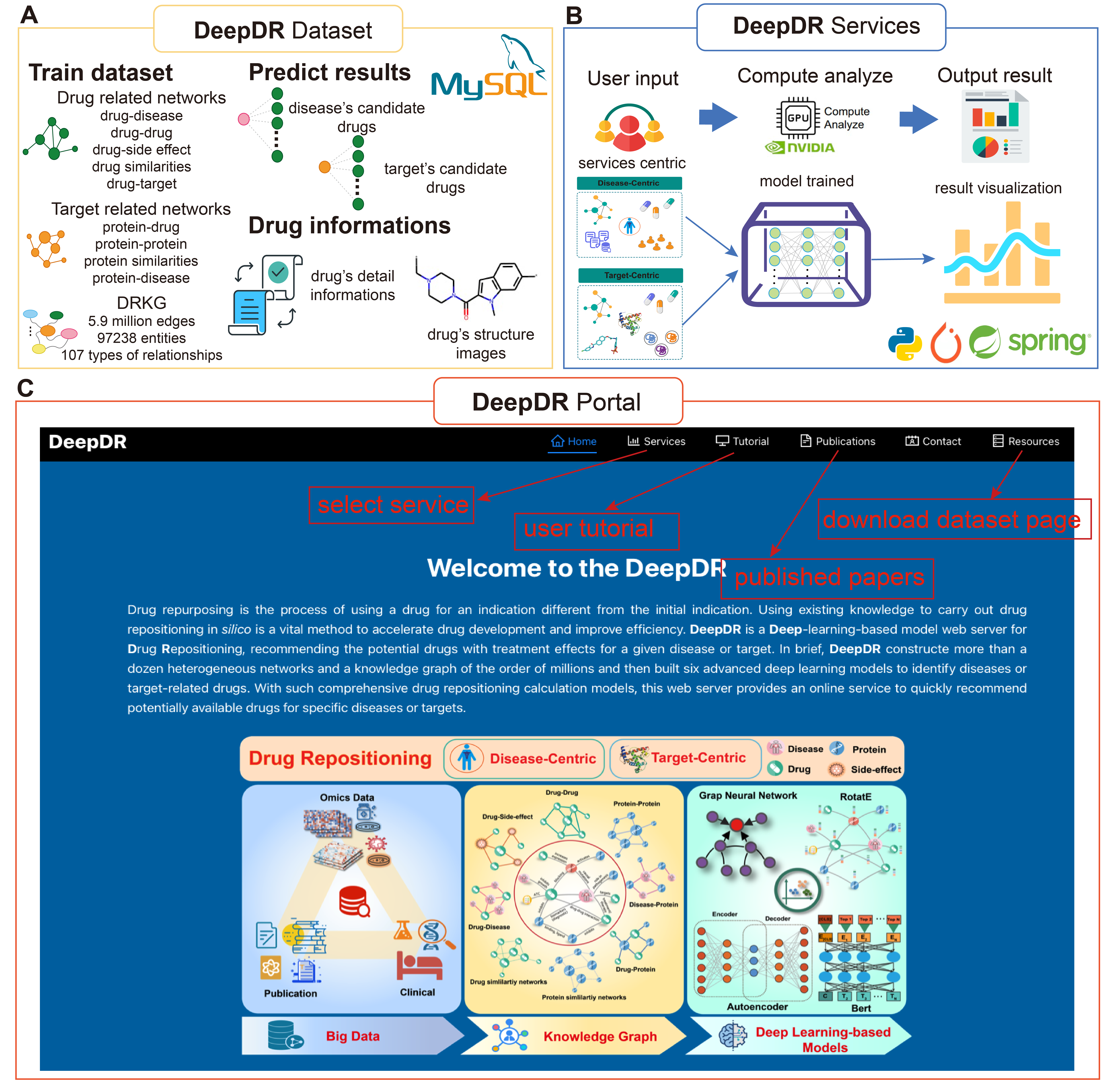}
\caption{\textbf{The overview of the DeepDR platform.} (A) {\bf DeepDR} dataset, (B){\bf DeepDR} services, and (C){\bf DeepDR} portal screenshot.}
\label{platform}
\end{figure*}

\begin{figure*}[h]
\centering
\includegraphics[scale=0.2]{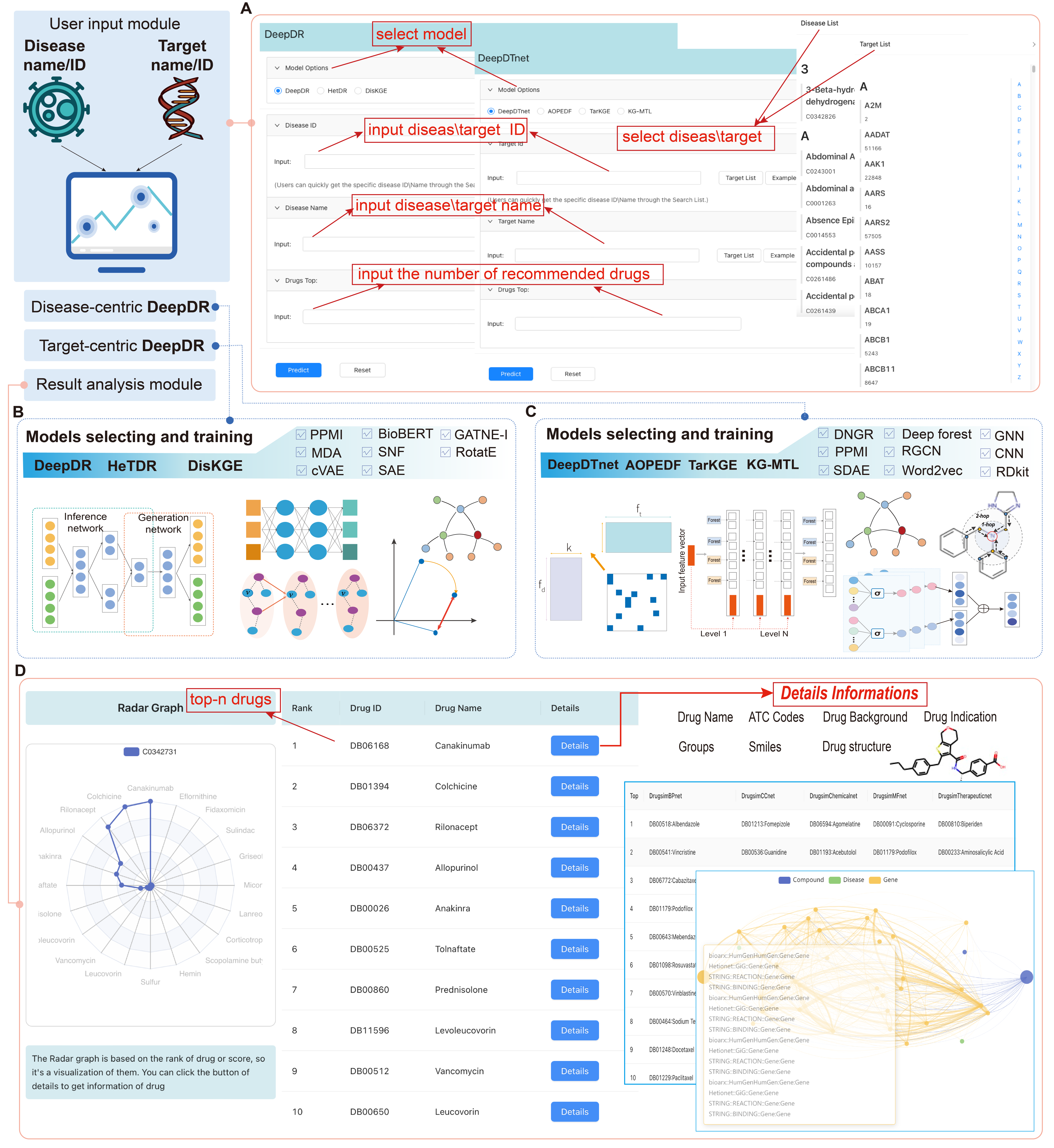}
\caption{\textbf{DeepDR services description.} (A) User input module, (B)disease-centric {\bf DeepDR} service module, (C) target-centric {\bf DeepDR} service module, and (D)result analysis module.}
\label{service}
\end{figure*}

{\bf DeepDR} (\href{http://drpredictor.com}{\textcolor{blue}{http://drpredictor.com}}), the first integrated online platform for drug repositioning, which fully automates the model training process. The system comprises three main components: {\bf DeepDR} Dataset, {\bf DeepDR} Services, and {\bf DeepDR} Portal, see Figure \ref{platform}. 
{\bf DeepDR} Dataset  (Figure \ref{platform}A) stores more than 15 networks and a comprehensive knowledge graph DRKG. DRKG contains 5.9 million edges across 107 types of relationships connecting drugs, diseases, proteins/genes, pathways, and expression from existing databases and a large scientific corpus of 24 million PubMed publications. These source data can be freely obtained and downloaded from the website platform. 
{\bf DeepDR} Services (Figure \ref{platform}B) integrate 6 deep learning tools developed by our research group (nearly 800 citations since the first release from 2019), including DeepDR \cite{zeng2019deepdr}, HetDR \cite{jin2021hetdr}, DeepDTnet \cite{zeng2020target}, AOPEDF \cite{zeng2020network}, Cov-KGE \cite{zeng2020repurpose}, and KG-MTL \cite{ma2022kg}, the details is shown in Table \ref{tab1}. For different prediction tasks, variants DisKGE and TarKGE of model Cov-KGE are proposed to predict candidate drugs for disease or target based on knowledge graphs, respectively. {\bf DeepDR} services include four components: (i)user input module (Figure \ref{service}A); (ii)disease-centric {\bf DeepDR} service module (Figure \ref{service}B); (iii) target-centric {\bf DeepDR} service module (Figure \ref{service}C); and (iv) result analysis module (Figure \ref{service}D). {\bf DeepDR} Portal serves as the web version of the system (Figure \ref{platform}C). It has a front end based on Python Django and HTML5, supported by the back end {\bf DeepDR} services. Upon entering the portal, users will see the home page, which mainly describes the basic information of the system and summarizes the highlights. The services page provides the two service centers for drug repositioning. The tutorial page provides a detailed overview of the {\bf DeepDR}’s operational route, components, and DL models. The contact page lists the contributors of the work and allows users to contact them via email for any assistance. The resources page allows authors to obtain all datasets for free.

\subsection{{\bf DeepDR} dataset}
The datasets can be summarized as follows: (i) DeepDR based on 10 networks: clinically reported drug–drug interactions, drug-disease interactions, drug–target interactions, drug-side-effect associations, and 6 drug similarity networks (chemical similarities, therapeutic similarities derived from the Anatomical Therapeutic Chemical Classification System, drugs’ target sequence similarities, Gene Ontology (GO) biological process, GO cellular component, and GO molecular function). (ii) HeTDR based on the above ten networks and a large scientific corpus of 24 million PubMed publications. (iii) DeepDTnet and AOPEOF based on 15 networks: clinically reported drug–drug interactions, drug-disease interactions, drug–target interactions, drug-side-effect associations, human protein–protein interactome, 6 drug similarity networks, and 4 proteins similarity networks (sequence similarities, Gene Ontology (GO) biological process, GO cellular component, and GO molecular function).
(iv) DisKGE, TarKGE, and KG-MTL based on a comprehensive biological knowledge graph called DRKG relating genes, compounds, diseases, biological processes, side effects, and symptoms. DRKG includes 5.9 million edges across 107 types of relationships connecting drugs, diseases, proteins/genes, pathways, and expression from a large scientific corpus of 24 million PubMed publications and six existing databases including DrugBank, Hetionet, GNBR, String, IntAct, and DGIdb.
\subsection{{\bf DeepDR} services and portal}
\subsubsection{User Inputs module}
The input module mainly includes the following parts (Figure \ref{service}A):
\begin{itemize}
\item 1. Model Options: the user can select a specific model for prediction.
\item 2. Disease or Target ID\textbackslash Name: If the user uses the ``Disease-Centric {\bf DeepDR}" service module, {\bf DeepDR} supports the input disease MeSH ID or Name. ``Disease List" provides all predictable diseases Name and ID to facilitate access to predictable disease information. And users can click on specific diseases to directly predict. ``Example" provides a sample example of automatic filling (e.g., C0342731/Deficiency of mevalonate kinase/20), users can directly click to view the predictive case.  If the user uses the ``Target-Centric {\bf DeepDR}” service module, {\bf DeepDR} supports the input target unique ID\textbackslash Name. ``Target List" provides all predictable targets Name and ID to facilitate access to predictable target information. "Example" provides a sample example of the automatic filling (e.g., 9971/NR1H4/20), users can directly click to view the predictive case.
\item 3. Drugs\_Top: Here users can choose to display the number of recommended drugs, and the web server will display the ranking of the corresponding number of drugs according to the predicted correlation.
\item 4. Operation buttons: After inputting the available disease\textbackslash target ID\textbackslash Name and Drugs Top, the user can select ``Predict" to make a prediction, and click ``Reset" to re-input other diseases.
\end{itemize}
The {\bf DeepDR} server is an automatic prediction and analysis platform, users can do this by either entering the genes or diseases manually into a box or choosing the gene or disease they are interested in from our given list of dictionary indexes. In addition, users can select different deep-learning models for prediction. 

\subsubsection{Prediction module}
{\bf DeepDR} aims to recommend the potential drugs for a given target or disease, based on six deep learning models developed by our group, including DeepDR, HetDR, DeepDTnet, AOPEDF, Cov-KGE, and KG-MTL, the details are summarized in the Table \ref{tab1}. The ``Disease-Centric {\bf DeepDR}'' service gives three optional models, including a heterogeneous networks-based model, a heterogeneous networks and text mining-based model, and a knowledge graph-based model (see Method Section  \ref{Disease-Centric}). The user can select a model and input a disease ID or name to obtain the results of the recommended drugs with ranking. The ``Target-Centric {\bf DeepDR}'' service gives four optional models, including two heterogeneous networks-based models, a knowledge graph-based model and a model based on knowledge graph with molecular structure (see Section \ref{Target-Centric}). The user can select a model and input a target ID or name to obtain the rank of the recommended drugs. {\bf DeepDR} provides the DL models that are well-trained with large-scale biological data, the users can obtain the predicted drugs by entering an entity (i.e., gene or disease) and selecting an interesting model. When the ``Predict'' button is hit, the selected machine learning model will automatically start training without the need for the user to upload training data. 

\begin{table*}[!h]
\centering
\resizebox{1\textwidth}{!}{
\begin{tabular}{ccccc}
\toprule
Category of sever centric &Methods & Databases & Descriptors & Code URL \\
\midrule

\multirow{3}{*}{Disease-Centric {\bf DeepDR}}
    & DeepDR\cite{zeng2019deepdr} & \makecell[c]{1519 drug; 1229 diseases; 1025 targets(proteins);\\ 12904 side-effect; 6677 DDiIs; 290836 DDIs; \\6744 DTIs; 382041 DSIs; \\6 drug-drug similarities networks} & \makecell[c]{multi-modal deep autoencoder \\learn drug features base on\\ heterogeneous biological networks} & \makecell[c]{https://github.com\\/ChengF-Lab/deepDR} \\
    & HeTDR\cite{jin2021hetdr} & \makecell[c]{the DeepDR databases and \\text corpora (English Wikipedia,\\ BooksCorpus, PubMed Abstracts,\\ and, PMC Full-text articles)}& \makecell[c]{base on the biomedical \\text and heterogeneous biological \\network DL method} & \makecell[c]{https://github.com/\\stjin-XMU/HeTDR} \\
    & DisKGE \cite{zeng2020repurpose} & DRKG\cite{ioannidis2020drkg}(97,238 entities, 5,874,261 edges, 13 entity-types)& \makecell[c]{an integrative, knowledge\\ graph-based  DL method}  & \makecell[c]{https://github.com/\\ChengF-Lab/CoV-KGE} \\

\midrule

\multirow{4}{*}{Target-Centric {\bf DeepDR}}
    & DeepDTnet\cite{zeng2020target} & \makecell[c]{732 drug;  440 diseases; 1915 targets(proteins);\\ 12904 side-effect; 1208 DDiIs; 132768 DDIs; \\5680 DTIs; 16133 PPIs; 263805 DSIs\\6 drug similarities networks;\\4 target similarities networks} & \makecell[c]{base on heterogeneous biological \\ networks DL method} & \makecell[c]{https://github.com/\\ChengF-Lab/deepDTnet}\\
    & AOPEDF\cite{zeng2020network} & same as the DeepDTnet databases & \makecell[c]{an arbitrary-order proximity \\ embedded deep forest approach base \\on heterogeneous biological networks} & \makecell[c]{https://github.com/\\ChengF-Lab/AOPEDF}\\
    & TarKGE \cite{zeng2020repurpose} & DRKG\cite{ioannidis2020drkg}(97,238 entities, 5,874,261 edges, 13 entity-types) &\makecell[c]{an integrative, knowledge\\ graph-based  DL method} & \makecell[c]{https://github.com/\\ChengF-Lab/CoV-KGE}\\
    & KG-MTL\cite{ma2022kg} & \makecell[c]{DRKG\cite{ioannidis2020drkg}(97,238 entities, 5,874,261 edges, 13 entity-types);\\DrugBank\cite{wishart2018drugbank} (5996 drug, 3479 target, 16553 DTIs);\\DrugCentral\cite{ursu2016drugcentral}(1427 drug, 1106 target, 9477 DTIs);\\human\cite{liu2015improving}(1080 compound, 816 proteins(target), 2471 CPIs); \\C.elegans\cite{liu2015improving}(886 compound, 806 proteins(target), 2547 CPIs)} & \makecell[c]{a large-scale Knowledge Graph \\enhanced Multi-Task Learning model\\ base on knowledge graph and molecular \\graph DL method}& \makecell[c]{https://github.com/\\xzenglab/KG-MTL}\\
\bottomrule
\end{tabular}
}
\caption{Summary of the models that built on the {\bf DeepDR} website web services.}
\label{tab1}
\end{table*}

\subsubsection{Result analysis and visualization}

\begin{figure}[h]
\centering
\includegraphics[scale=0.6]{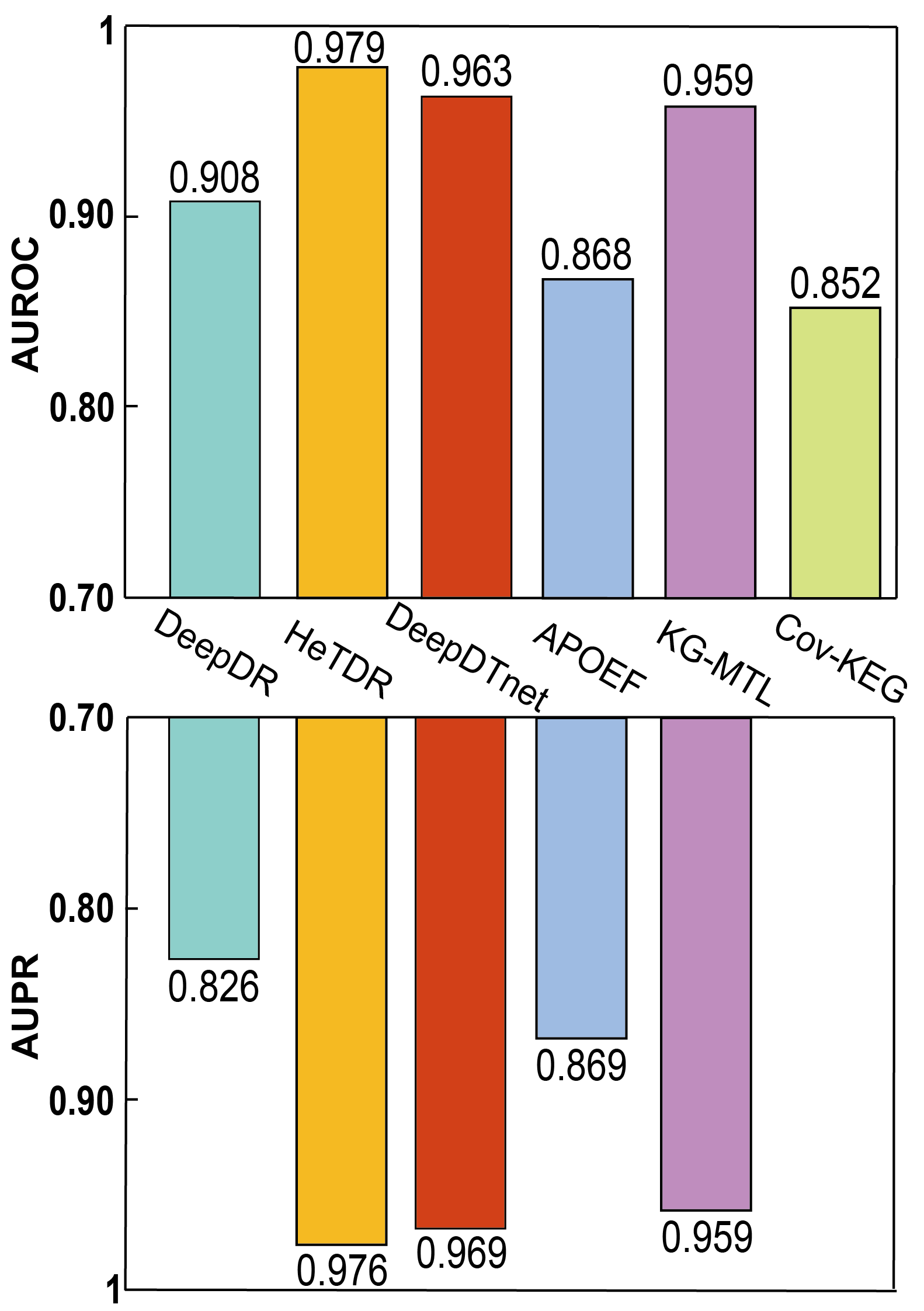}
\caption{\textbf{Performance of models built on {\bf DeepDR} web-server.} }
\label{AUROC}
\end{figure}

Recommended drug results can be found on the ``Results'' page, which is sorted by the degree of predicted relevance, and shows the ID and Name of the top-n drugs that are recommended to the users (Figure \ref{service}D).
It is worth noting that for each predicted drug, the user can click the corresponding ``Detail'' button to jump to the details page, where all the information of the drug is displayed, including ATC Codes, Drug Background, Drug Indication, Drug structure, and so on. Figure \ref{AUROC} illustrates the performance of the models on the {\bf DeepDR} web-server platform, which is derived from the performance evaluation in the published works of each model. 
It can be seen from the figure that the models implemented on the website have achieved excellent performance in their respective tasks. Users can review the summary and introduction of each model in section \ref{method} and select the appropriate model based on their specific requirements for drug repositioning prediction. 

In addition, to assist users in better understanding why the model predicts related drugs, for the network-based model, the most relevant top 20 drugs are given in five relationships of the drug in therapeutic, chemical, Gene Ontology (GO) biological process, GO cellular component and GO molecular function. For the knowledge graph-based model, a network visualization of the paths between the drug and the gene or disease that the user wants to predict is provided. The user can interact with the network by clicking on a node or edge to view the corresponding entity or relationship type it represents.




\section{Discussion and conclusion}
Although the implementation of DL-based methods in computer-aided drug discovery is accelerating at an unprecedented rate, the experience required in training and deploying DL models limits the applicability for medicinal chemists and bioinformaticians. Therefore, a user-friendly online platform is urgently needed, especially for researchers without a computer science background. In this study, we establish {\bf DeepDR}, which is the first integrated deep learning-based platform for drug repositioning. {\bf DeepDR} integrates six advanced deep learning-based models, which have been successfully published in well-known journals. 
These models can accurately predict potential drugs for known diseases or targets, and users can choose specific models according to the desired scenarios. In addition, the web server shows the relevant details of predicted candidate drugs and provides a visual analysis derived from the trained KG.
In response to the rapidly evolving drug discovery data, we consistently update our web server's database to offer the latest and most relevant information. This commitment not only bolsters prediction accuracy but also provides users with up-to-date insights. In the future, we're expanding our collection of drug discovery models and enhancing prediction speed.

In summary, {\bf DeepDR} are powerful tools that can accelerate the drug discovery process by harnessing the capabilities of high-performance computing and DL algorithms.  It also enables collaboration between researchers and pharmaceutical companies by facilitating the sharing of data, resources, and expertise.

\section{METHODS}
\label{method}
\subsection{Web-server implementation}
DeepDR utilizes a deep-learning model optimized for high-performance GPU architectures and a fast CPU-backed database. It runs on Ubuntu 18.04 Linux, with multiple Intel Xeon Silver 4210R CPUs and 256 GB RAM. The model processes data using vast SSD storage and NVIDIA RTX 3090 GPU clusters. The platform's interface is built with the React framework, backend through Spring Boot, and data storage via MySQL. ECharts aids in creating interactive visualizations for user-friendly results display. DeepDR is compatible with major browsers, including Internet Explorer ($\geqslant$v.7.0), Firefox, Edge, Safari, and Chrome. In conclusion, with our DeepDR platform, experimental and computational scientists can easily obtain comprehensive prediction analysis based on the output data from these deep-learning models through an easy-to-use, systematic, highly accurate, and computationally automated platform.

\subsection{Overview and Guidance of Disease-Centric Models}
\label{Disease-Centric}
The "Disease-Centric {\bf DeepDR}" service mainly recommends the most effective drugs to treat the disease inputted by the user.
In the module, {\bf DeepDR} gives three optional models, including a heterogeneous network-based model (DeepDR), a heterogeneous network and text mining-based model (HeTDR), and a knowledge graph-based model (DisKGE). Here we summarize the calculation schemes and suitable task requirements of these three DL models.
\subsubsection{DeepDR}
DeepDR first uses random walk-based representation as a pre-processing step to obtain the PPMI matrices $X^i \in \mathbb{R}^{n \times n}, i\in{1,2,\dots,N}$ of $n$ drugs, which can mitigate the sparsity of some individual network types, $N$ is the number of networks. Then, DeepDR integrates $N$ networks represented by PPMI matrices to obtain drug common features $X$ using MDA. The low-dimensional and high-level common features $X$ of the drug are extracted from the hidden layer of MDA, the MDA optimizes the drug features by minimizing the reconstruction loss between each original and reconstructed PPMI matrix, the loss defined as follows:
\begin{equation}
\sum \limits_{i=1}^{N} Loss(X^i, \hat{X}^i)
\end{equation}
Then the learned low-dimensional representation of drugs together with clinically reported drug-disease pairs are encoded and decoded collectively via a collective Variational Autoencoder(cVAE), to infer candidates for approved drugs for which they were not originally approved. The drug-disease pairs are defined as $Y$,  the cVAE to generalize the linear models for top-$N$ recommendation with side information to non-linear models. The $X$, $Y$ are recovered through:
\begin{equation}
\left\{
    \begin{array}{ll}
    U \sim f_\phi(Y), Y \sim f_\theta(U)\\ 
    Z \sim f_\phi(X), X \sim f_\theta(Z)
    \end{array}
\right.
\end{equation}
where $U$ is the matrix of latent drug representation, $Z$ is the matrix of latent feature representation, $f_\phi(\dot)$ and $f_\theta(\dot)$ correspond to the inference network and generation network of cVAE parameterized by $\phi$ and $\theta$, respectively. This model focuses on obtaining feature information of drugs from biological networks for drug repositioning.

\subsubsection{HeTDR}
HeTDR uses SNF to integrate nine drug-related networks into one network with global information and then utilizes the SAE to obtain the high-quality representation of the drug features. Secondly, HeTDR uses the BioBERT model to obtain the feature information of disease from biomedical corpora. Specifically, HeTDR uses the pre-trained parameters of the BioBERT and selects the relation extraction task for fine-tuning training. After the fine-tuning process, HeTDR extracts the representation of sub-words and obtains the representations of all diseases by frequent subwords. Finally, HeTDR combines the drug-disease association network with the features information of the drug and disease to infer the potential associations between drug and disease. The $l$-th layer embedding of node $v_i$ is aggregated from the neighbor’s embedding. The neighborhood aggregation embedding is computed as:
\begin{equation}
\mathbf{n}_i^{l}=\sigma({\widehat{\boldsymbol{W}}}^{l}\cdot MEAN(\left\{\mathbf{n}_j^{l-1},\forall v_j\in\mathcal{N}_i\right\}))
\end{equation}
where $\sigma$ is an activation function, $\widehat{\boldsymbol{W}}^{l}$ is the weight matrices to propagate information between different layers, and $\mathcal{N}_i$ are the neighbors set of node $v_i$. The initial neighborhood aggregation embedding $\mathbf{n}_i^{0}=g_t(\mathbf{x}_i)$, $\mathbf{x}_i$ is the features information of node $v_i$.
The subscript $t$ is $v_i$’s corresponding node type and $g_t$ is a transformation function.
In order to preserve the spatial approximation of node attributes and the topological structure information of the drug-disease association network, the final drug-disease association prediction is performed using the attribute heterogeneous graph neural network algorithm of node attribute embedding, neighborhood set embedding, and base embedding. The overall embedding of node $v_i$'s function is as follows:
\begin{equation}
\left\{
    \begin{array}{ll}
    \mathbf{v}_i=b_t(\mathbf{x}_i)+\alpha \boldsymbol{M}^T \mathbf{n}_i R_i+\beta \boldsymbol{D}_t^T(\mathbf{x}_i)\\ 
    R_i=softmax(\boldsymbol{w}^Ttanh(\boldsymbol{W}n_i))^T
    \end{array}
\right.
\end{equation}
where $\alpha$ is a hyperparameter, $\boldsymbol{M}\in\mathbb{R}^{f_n\times f_o}$ is a trainable transformation matrix, $f_n$ is the dimension of neighborhood aggregation embedding, $f_o$ is the dimension of overall embedding, $\boldsymbol{w}$ is the trainable parameter with size $d_a$, $\boldsymbol{W}$ is the trainable parameter with size $d_a\times f_n$, $\beta$ is a coefficient, the symbol $T$ represents the transposition of the matrix or the vector, and $\boldsymbol{D}_t$ is a feature transformation matrix on $v_i$'s corresponding node type $t$. Then, HeTDR uses meta-path-based random walks and skip-gram to learn drug and disease embeddings based on drug-disease heterogeneous networks. The probability of node $v_j$ given $v_i$ is defined as:
\begin{equation}
\label{2-20}
P_\varepsilon(v_j\vert v_i)=\frac {exp{(c_j^T\cdot \mathbf{v}_i)}}{\sum_{k\in V_t}{exp{(c_k^T\cdot \mathbf{v}_i)}}}
\end{equation}
where all parameters are defined as $\varepsilon$, $c_k$ is the context embedding of node $v_i$ and $\mathbf{v_i}$ is the overall embedding of node $v_i$. This model is more suitable for drug repositioning prediction that requires additional consideration of semantic relationships between diseases. 
\subsubsection{DisKGE}
DisKGE is a variant from CovKGE,  which applies a deep learning methodology, RotatE \cite{sun2019rotate}, for \emph{in silico} prediction of the small molecular drugs for known diseases, from knowledge graph DRKG \cite{ioannidis2020drkg}. The design of the RotatE method is inspired by Euler's identity $e^{i\theta} = cos\theta+i sin\theta$.  RotatE model maps entities and relations into a complex vector space and defines each relation as a rotation from a source entity to a target entity. Specifically, given a triplet $(h,r,t)$ from the knowledge graph, for each dimension in the complex space, RotatE expects that $\mathbf{t} = \mathbf{h}\circ\mathbf{r}$, where $\mathbf{h},\mathbf{r},\mathbf{t} \in \mathbb C$ are the embedding of entities $h,t$ and relation $r$, modulus $\vert r_i\vert =1$, and $\circ$ denotes the Hadamard (element-wise) product. The distance function of RotatE for each triple is as follows:
\begin{equation}
d_r(\mathbf{h},\mathbf{t})=\Vert \mathbf{h}\circ\mathbf{r}-\mathbf{t}\Vert
\end{equation}
DisKGE leveraged RotatE that an unsupervised knowledge graph embedding method to learn low-dimensional but informative vector representations for all entities (e.g., drugs, diseases) in DRKG. Subsequently, DisKGE predicts an initial list of drug candidates that show potential for the treatment of each disease in DRKG by computing the distance between them. This model is more suitable for drug repositioning prediction that considers the global knowledge comprehensively, and there are certain interpretability requirements for predicting results.

\subsection{Overview and Guidance of Target-Centric models}
\label{Target-Centric}
The "Target-Centric {\bf DeepDR}" service mainly recommends the most likely drug related to the gene inputted by the user. In the module, {\bf DeepDR} give four optional models, including two heterogeneous networks-based models (DeepDTnet and AOPEOF), a knowledge graph-based model (DisKGE), and a knowledge graph-based model with molecular structure graph (KG-MTL). Here we summarize the calculation schemes and suitable task requirements of these four deep models.
\subsubsection{DeepDTnet}
DeepDTnet is a deep learning methodology for new target identification and drug repurposing in a heterogeneous drug-gene-disease network embedding 15 types of chemical, genomic, phenotypic, and cellular network profiles. DeepDTnet first uses the DNGR \cite{cao2016deep} embedding model to learn features. DNGR model uses random surfing-based representation as a pre-processing step prior to obtaining the PPMI matrices $X \in \mathbb{R}^{n \times n}$ of $n$ drugs, which can mitigate the sparsity of some individual network types. Then the learned low-dimensional and high-level features $\mathbf{x}$ of each vertex from the different network's PPMI matrices by a stacked denoising autoencoder (SDAE).  A SDAE model minimizes the regularized problem and tackles reconstruction error, defined as follows:
\begin{equation}
\min \limits_{\mathbf{W}_l,\mathbf{B}_l}\Vert \mathbf{x}-\widehat{\mathbf{x}}\Vert _F^2+\lambda\sum\limits_{l}\Vert \mathbf{W}_l \Vert _F^2
\end{equation}
where $\mathbf{W}_l$ and $\mathbf{B}$ is weight matrix and bias vector of layer $l$ respectively, $\Vert \bullet \Vert_F$ is the Frobenius norm and $\lambda$ is a regularization parameter.  The representations of vertices are extracted from the middle layer of SDAE. 

And then owing to the lack of experimentally reported negative samples from the publicly available databases, DeepDTnet employs the Positive Unlabeled (PU)-matrix completion algorithm for low-rank matrix completion. The optimization problem of DeepDTnet is parameterized as:
\begin{equation}
\begin{aligned}
\min \limits_{\mathbf{P},\mathbf{O}} &\sum\limits_{(i,j)\in \Theta^+}(\mathbf{M}_{ij}-\mathbf{x}_i^d \mathbf{P} \mathbf{O}^T (\mathbf{x}_j^t)^T)^2+\epsilon\sum \limits_{(i,j)\in \Theta^-}\\ &(\mathbf{M}_{ij}-\mathbf{x}_i^d \mathbf{P} \mathbf{O}^T (\mathbf{x}_j^t)^T)^2 + \lambda(\Vert \mathbf{P} \Vert ^2_F + \Vert \mathbf{O} \Vert ^2_F)
\end{aligned}
\end{equation}
where $\mathbf{x}_i^d \in \mathbb{R}^{f_d}$ is the feature vector of drug $i$, $\mathbf{x}_j^t \in \mathbb{R}^{f_t}$ is the feature vector of target $j$, $P\in \mathbb{R}^{f_d \times k}$ and $O\in \mathbb{R}^{f_t \times k}$ share a low dimensional latent space, $M \in \mathbb{R}^{ n_d \times n_t}$ is the drug-target interactions matrix, $f_d$ and $f_t$ is the feature dimensions of drugs and targets, $n_d$ and $n_t$ is the number of drugs and targets, respectively, $k<<n_d,n_t$. The set $\Theta^+$ is the known interactions, $\Theta^-$ is the unobserved interactions, and the $\epsilon<1$ is a key parameter, which determines the penalty of the unobserved entries toward $0$. Finally, the interaction score of drug $i$ and target $j$ can be approximated as:
\begin{equation}
S(i,j)=\mathbf{x}_i^d \mathbf{P} \mathbf{O}^T (\mathbf{x}_j^t)^T
\end{equation}
The model is a powerful network-based deep learning methodology for target identification to accelerate drug repurposing and the results predicted have been verified by biological experiments.

\subsubsection{AOPEOF}
AOPEOF is a network-based computational framework, which uses an arbitrary-order proximity to preserve network embedding and deep forest to predict drug-target interactions. AOPEOF first preserves complementary order proximity information for 15 heterogeneous networks by arbitrary-order proximity preserved network embedding. The high-order proximity is defined as a polynomial function $\mathcal{F}$ of adjacency matrix $\mathbf{M}$:
\begin{equation}
\mathbf{S}=\mathcal{F}(\mathbf{M}) = w_1\mathbf{M}+w_2\mathbf{M}^2+\dots+w_l\mathbf{M}^l
\end{equation}
where $w_1,\dots,w_l$ is the weight, $l$ is the order, and $w_i>0, 0<i<l$. When the summation converges, the $l$ is allowed to $+\infty$. 
Matrix factorization is a widely adopted method to preserve the high-order proximity in a low-dimensional vector space, which minimizes the following objective function:
\begin{equation}
\min\limits_{\mathbf{U^*}\mathbf{V^*}} \Vert \mathcal{S} -\mathbf{U^*}\mathbf{V^*}^T \Vert_F^2
\end{equation}
where $\mathbf{U^*}, \mathbf{V^*} \in \mathbb{R}^{n\times f_s}$ are context embedding vectors, $n$ is the number of nodes in the network, $f_s$ is the dimensionality of the space. Without loss of generality, AOPEDF uses $\mathbf{U^*}$ as the content embedding vectors. From the Ercart-Young theorem, the global optimal solution to the above objective function can be obtained by truncated SVD \cite{eckart1936approximation}. In order to save computing time and consumption, and to solve problems where different networks and target applications usually require proximities of different orders. AOPEDF transforms the SVD problem in the above objective function into an eigen-decomposition problem \cite{zhang2018arbitrary}. After learning the low-dimensional vector representation of drugs and proteins, AOPEDF utilizes the deep forest for DTIs prediction.
In total, the model constructed a heterogeneous network by uniquely integrating 15 networks covering chemical, genomic, phenotypic, and network profiles among drugs, proteins/targets, and diseases. Then, AOPEDF builds a cascade deep forest classifier to infer new drug-target interactions.

\subsubsection{TarKGE}
Similar to DisKGE, TarKGE applies RotatE that is an unsupervised knowledge graph embedding method to learn low-dimensional but informative vector representations for all entities (e.g., drugs, diseases, targets) in DRKG. Subsequently, TarKGE predicts a list of drug candidates that show potential for interaction of each target in KG through computing the distance between them by a score function. After getting the list of drug candidates, TarKGE selects the top 20 drugs for every target in KG. This model also is more suitable for drug-target prediction that considers the global knowledge comprehensively, and there are certain interpretability requirements for predicting results.
\subsubsection{KG-MTL}
KG-MTL \cite{ma2022kg} is a large-scale knowledge graph enhanced multi-task Learning model, called KG-MTL, which extracts the features from both the knowledge graph and molecular structure graph in a synergistic way. The model is the first work to apply a large-scale knowledge graph on a multi-task learning model and consists of three major modules. Specifically, (i) the DTI module is used to extract the features of drugs and related entities from large-scale KG (DRKG). In this module, KG-MTL employs a 3-layer RGCN model \cite{thanapalasingam2022relational} to extract the semantic relations and topological structure of entities from the subgraph, defined as follows:
\begin{equation}
\mathbf{x}_{i}^{l}=\sigma(\sum\limits_{r\in\mathcal{R}}\sum \limits_{j\in\mathcal{N}_i^r} \frac{1}{\vert \mathcal{N}_i^r \vert} \mathbf{W}_r^{l-1}\mathbf{x}_j^{l-1}+\mathbf{W}_o^{l-1}\mathbf{x}_i^{l-1})
\end{equation}
where $\mathbf{x}^l$ is the embedding of entity $i$ in the $l$-th layer, $\sigma$ is the activation function, $r$ is the relation type, $\mathcal{R}$ is the set of relations, $\mathcal{N}_i^r$ denotes the neighbors of entity $i$ under relation $r$, $\mathbf{W}_r^l$ is the relation-specific weight for relation $r$ in the $l$-th layer, $\mathbf{W}_o^l$ is the a self-loop weight for entity $i$ in the $l$-th layer. The module can obtain the embedding $\mathbf{x}_{d_i}$ of drug entities and the embedding $\mathbf{x}_{t_i}$ of target entities, and then DTI probability is calculated using a multi-layer perception (MLP).

(ii) the CPI module uses CNN to learn protein sequence representations, and adopts GCN to learn drug molecular graph representations. The drug molecular graph representations can be denoted as:
\begin{equation}
\mathbf{x}_{g_i}=\frac{1}{V}\sum\limits_{a=1}^{\vert V \vert}\sigma(f(\mathbf{v}_a))
\end{equation}
where $\mathbf{x}_{g_i}$ is the embedding of molecule graph $i$, $V$ is the number of atoms in the molecule graph, $v_a\in V$ is the $a$-th atom, which is initialized by a 78-dimensional feature vector $\mathbf{v}_a$, $f(\dot)$ is the message passing function.

(iii) Shared Unit is designed to share task-independent drug features between the previous two modules by combining the drug molecular representation of the compound and corresponding drug entity embedding from KG. The shared unit first uses four trainable weights $(\mathbf{W}_{dd},\mathbf{W}_{dg},\mathbf{W}_{gg},\mathbf{W}_{gd})$ to automatically learn and reconstruct the drug and compound feature by linear transformation as follow:
\begin{equation}
\left\{
    \begin{array}{ll}
    \mathbf{x}'_d = \mathbf{W}_{dd}^T \odot \mathbf{x}_{d_i}^{l}+ \mathbf{W}_{gd}^T\odot \mathbf{x}_{g_i}\\ 
    \mathbf{x}'_g = \mathbf{W}_{gg}^T \odot \mathbf{x}_{g_i}+ \mathbf{W}_{dg}^T\odot \mathbf{x}_{d_i}^l
    \end{array}
\right.
\end{equation}
where $\odot$ is the element-wise multiplication. Then shared unit construct a cross matrix $\mathcal{C}$ by pairwise interactions of $\mathbf{x}'_d$ and $\mathbf{x}'_g$ as follows:
\begin{equation}
\mathcal{C} = \mathbf{x}'_d(\mathbf{x}'_g)^T
\end{equation}
Finally, the shared unit uses a non-linear operator to project them back to the original feature space of the input of DTI and CPI modules, and the non-linear operator is calculated as follows:
\begin{equation}
\left\{
    \begin{array}{ll}
    \mathbf{x}''_{d_i} = \mathcal{C}\otimes \mathbf{W}'_dd + \mathcal{C}^T \otimes \mathbf{W}'_{gd} + \mathbf{b}_d\\ 
    \mathbf{x}''_{g_i} = \mathcal{C}\otimes\mathbf{ W}'_gg + \mathcal{C}^T \otimes \mathbf{W}'_{dg} + \mathbf{b}_g
    \end{array}
\right.
\end{equation}
where $\otimes$ is the matrix multiplication, $(\mathbf{W}'_{dd},\mathbf{W}'_{dg},\mathbf{W}'_{gg},\mathbf{W}'_{gd})$ are trainable weights, $\mathbf{b}_d$ and $\mathbf{b}_g$ is the bias vector. The $\mathbf{x}''_{d_i}$ will be input to the $l+1$-th RGCN layer of the DTI module, the $\mathbf{x}''_{g_i}$ will be input to the CPI module to obtain a new embedding by a DNN.
This model is more suitable for considering multiple types of relationships, as well as drug repositioning predictions where certain targets are more sensitive to drug structural information.

\section*{Key points}
\begin{itemize}
    \item The {\bf DeepDR} platform is proposed for the prediction of the potential drugs for a given disease or target, which is a pioneering integrated online platform for drug repositioning. 
    \item The {\bf DeepDR} platform fully automates the model training process and integrates 6 deep learning models developed by our research group. These models have undergone rigorous peer review and have been published in high-level journals, accumulating nearly 800 citations since their initial release in 2019.
    \item The {\bf DeepDR} platform explores the relationship between predicted drugs and a given disease or target, and provides the visualization analysis and relevant details for each recommended drug, including drug background, indications, molecular structure, similar top-20 drugs, and so on.
    \item The new servers have been running for over one year and have received nearly 8,000 visits. For {\bf DeepDR}, the "Disease-Centric {\bf DeepDR}" and the "Target-Centric {\bf DeepDR}" service both can immediately give predictive results.
    \item {\bf DeepDR} is open source and free to use.
\end{itemize}

\section*{Availability and requirements}
\begin{itemize}
     \item \textbf{Project name:} \textbf{DeepDR}.
     \item \textbf{Project home page:} \href{https://drpredictor.com}{\textcolor{blue}{https://drpredictor.com}}.
     \item \textbf{Operating system(s):} Linux.
     \item \textbf{Programming language:} Python, Java, javascript.
     \item \textbf{Other requirements:} see \href{https://github.com/stjin-XMU/DeeDR_web-server}{\textcolor{blue}{https://github.com/stjin-XMU/DeeDR\_web-server}}.
     \item \textbf{License:} GNU GPL.
     \item \textbf{Any restrictions to use by non-academics:} see licence.
\end{itemize}

\section*{Availability of data and materials}
The source code and datastes are available at: \href{https://github.com/stjin-XMU/DeeDR_web-server}{\textcolor{blue}{https://github.com/stjin-XMU/DeeDR\_web-server}}.

\section*{Ethics approval and consent to participate}
Not applicable.

\section*{Consent for publication}
Not applicable.

\section*{Competing interests}
The authors declare no competing interests.

\bibliography{sn-bib}


\end{document}